\documentclass{elsarticle}

\usepackage{lineno,hyperref}
\usepackage[usenames, dvipsnames]{xcolor}
\usepackage{soul}
\usepackage{amsmath}
\usepackage{float}
\usepackage{subcaption}
\usepackage{natbib}
\setcitestyle{authoryear,open={(},close={)}}
\usepackage{adjustbox}
\definecolor{sand}{RGB}{255,255,187}
\definecolor{shale}{RGB}{142,142,139}

\usepackage{ulem}
 
\newcommand{\WOB}{\text{WOB}}
\newcommand{\RPM}{\text{RPM}}
\newcommand{\TRQ}{\text{TRQ}}
\newcommand{\Qin}{\text{Q in}}
\newcommand{\Qout}{\text{Q out}}
\newcommand{\SPP}{\text{SPP}}
\newcommand{\ROP}{\text{ROP}}
\newcommand{\HL}{\text{HL}}
\newcommand{\APR}{\text{APR}}
\newcommand{\SED}{\text{SED}}

\journal{Journal of Petroleum Science and Engineering}

\begin{document}

\begin{frontmatter}
 
\title{Data-driven model for the identification of the rock type at a drilling bit}

\author[SKT_address]{Nikita Klyuchnikov\fnref{design,implementation,drafting,analysis}}
\cortext[mycorrespondingauthor]{Corresponding author}
\ead{nikita.klyuchnikov@skolkovotech.ru}
\author[SKT_address]{Alexey Zaytsev\fnref{design,implementation,drafting}}
\author[IBM_address]{Arseniy Gruzdev\fnref{data,implementation,drafting}}
\author[SKT_address]{Georgiy Ovchinnikov\fnref{design,implementation,drafting}}
\author[SKT_address]{Ksenia Antipova\fnref{design,analysis}}
\author[SKT_address]{Leyla Ismailova\fnref{literature_review,drafting}}
\author[SKT_address]{Ekaterina Muravleva\fnref{literature_review,drafting}}
\author[SKT_address]{Evgeny Burnaev\fnref{design}}
\author[IBM_address]{Artyom Semenikhin\fnref{design}}
\author[GPN_address]{Alexey Cherepanov\fnref{data,revising}}
\author[GPN_address]{Vitaliy Koryabkin\fnref{design,revising}}
\author[GPN_address]{Igor Simon\fnref{design}\fnref{data,revising}}
\author[GPN_address]{Alexey Tsurgan\fnref{design,revising}}
\author[GPN_address]{Fedor Krasnov\fnref{data}}
\author[SKT_address]{Dmitry Koroteev\fnref{design,drafting,analysis}}

\fntext[design]{Conception and design of study}
\fntext[implementation]{Implementation of methods}
\fntext[data]{Acquisition of data}
\fntext[analysis]{Analysis and interpretation of data}
\fntext[drafting]{Drafting the manuscript}
\fntext[literature_review]{Literature review}
\fntext[revising]{Revising the manuscript}

\address[SKT_address]{Skolkovo Institute of Science and Technology, Skolkovo Innovation Center, Building 3,
Moscow  143026, Russia}
\address[IBM_address]{IBM East Europe/Asia, 10, Presnenskaya emb., Moscow, 123112, Russia}
\address[GPN_address]{Gazprom Neft Science \& Technology Centre, 75-79 liter D Moika River emb., St. Petersburg 19000, Russia}

\begin{abstract}
Directional oil well drilling requires high precision of the wellbore positioning inside the productive area. However, due to specifics of engineering design, sensors that explicitly determine the type of the drilled rock are located farther than 15m from the drilling bit. As a result, the target area runaways can be detected only after this distance, which in turn, leads to a loss in well productivity and the risk of the need for an expensive re-boring operation. 

We present a novel approach for identifying rock type at the drilling bit based on machine learning classification methods and data mining on sensors readings. We compare various machine-learning algorithms, examine extra features coming from mathematical modeling of drilling mechanics, and show that the real-time rock type classification error can be reduced from 13.5\% to 9\%. The approach is applicable for precise directional drilling in relatively thin target intervals of complex shapes and generalizes appropriately to new wells that are different from the ones used for training the machine learning model.

\end{abstract}


\end{frontmatter}


\section{Introduction}

Oil and Gas reserves are becoming more complex for an efficient exploration with significant financial margins nowadays. There is a number of examples when oil companies have to approach thin oil/gas bearing layers of complex topology. These layers, or the target intervals, can be as thin as a couple of meters. One of the common ways of exploring such intervals is directional drilling. 

The directional drilling aims to place a wellbore in a way that it has the maximal contact with the thin target layer. Later requires the wellbore trajectory to follow all the folds of the layer as accurate as possible. To follow the folds, drilling engineers use Logging While Drilling (LWD) data recorded by physical sensors placed on a borehole assembly 15 m to 40 m behind the drilling bit. The sensor data is the source of information on whether the sensors are within the oil/gas bearing formation or not. Based on this information, engineers correct the drilling trajectory. 

\begin{figure}[H]
\centering
\begin{subfigure}[b]{.2\linewidth}
\includegraphics[width=1\textwidth]{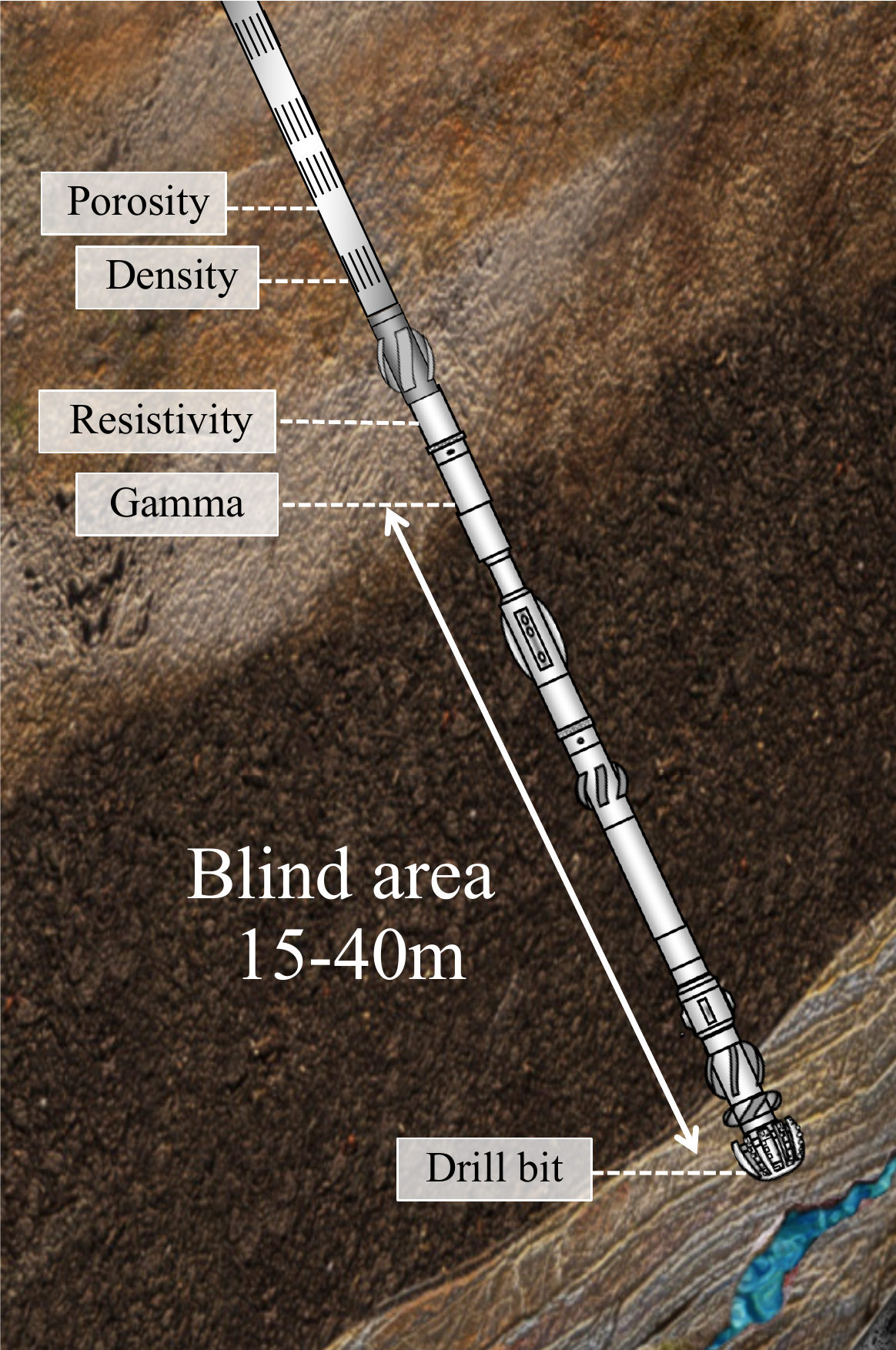}
\end{subfigure}
\begin{subfigure}[b]{.68\linewidth}
\includegraphics[width=1\textwidth]{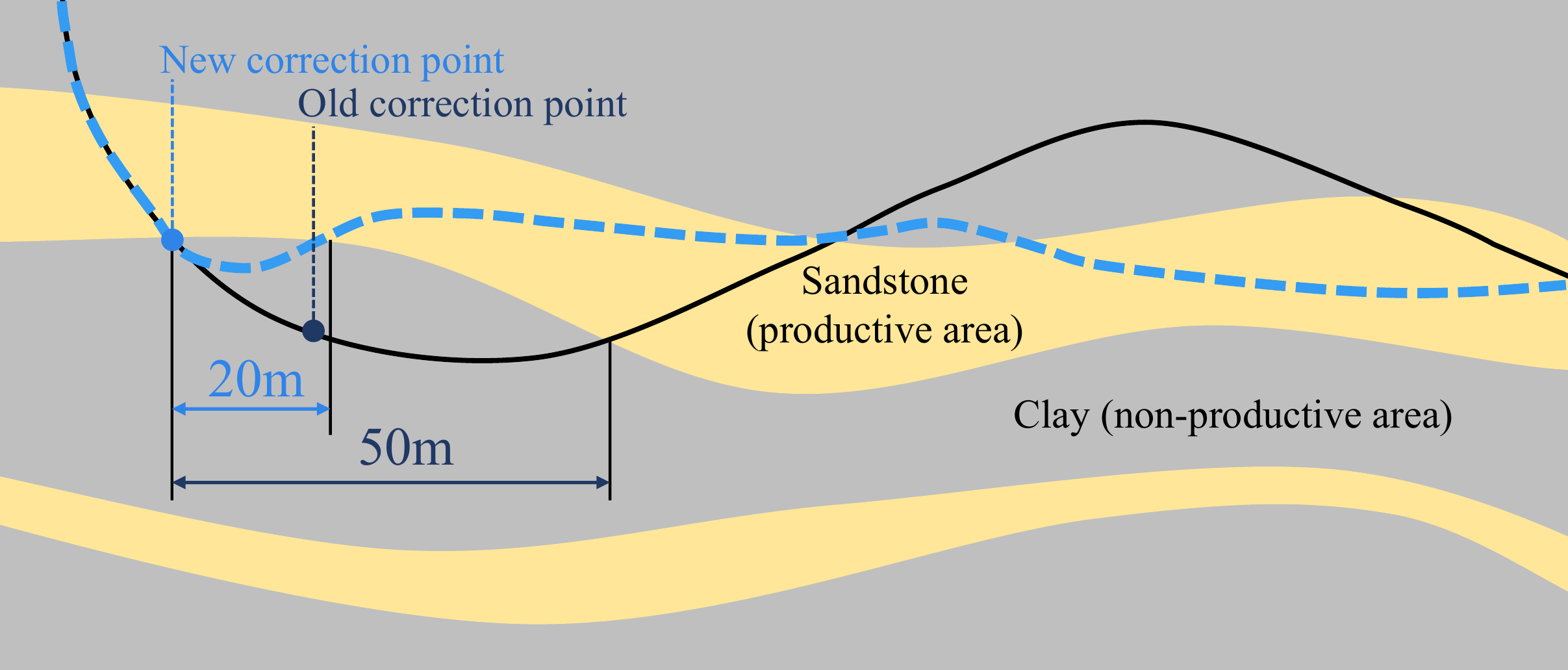}

\end{subfigure}
\caption{Schematic illustration of the drilling string (on the left) and the effect of timely applied trajectory correction (on the right): the black curve shows a trajectory in case rock types are available only at the distance of 15 m from the drilling bit, blue dashed curve corresponds to the trajectory when rock types are available at the drilling bit.}
\label{fig:drilling_schema}
\end{figure}

The gap between the bit and the sensors is a significant issue preventing the timely correction of the drilling trajectory. It can result in a non-optimal placement of the well or multiple cost-intensive re-drilling exercises. Figure \ref{fig:drilling_schema} shows schematic illustrations to supply the definition of the problem.   

This paper proves the feasibility of a technology aimed at optimizing trajectories of directional wells ensuring best possible contact of the wellbore and the target layer of the reservoir. The technology allows tackling the challenge of a delayed reaction on trajectory correction during drilling of directional wells. Machine learning allows eliminating $15$ m to $40$ m gap between the drilling bit and the LWD sensors and corresponding speeding up of decision making at the trajectory correction. 
Along with machine learning approaches we examine, 
how mathematical modeling can advance machine-learning based approaches.

Basically, a trained data-driven algorithm allows a computer to identify when the bit touches a shale-rich part of the formation by a continuous screening through the real-time Measurements While Drilling (MWD) data. 
In machine learning, this problem is referred as the two-class classification problem: we need to create a predictive classification model (a classifier) that can identify whether the bit at the current moment is in the shale-rich part of the formation (the first class) or not (the second class). In addition to labeling objects, the classifier can output the probability of the object to belong to a specific class, thus allowing to introduce confidence of predictions. 

From the machine learning perspective, the main peculiarities of the problem are missing values in measurements and a relatively high imbalance of classes: there are only 13.5\% of shales and hard-rocks in the available data, where "hard" refers to a measure of the resistance to localized plastic deformation induced by either mechanical indentation or abrasion, and 86.5\% of sands. Therefore, we tested different machine learning algorithms under these peculiarities, and developed appropriate evaluation methods of their performance.

The main contribution of this work is a novel data-driven approach for identifying lithotype at the drilling bit. We prove the feasibility of this approach by studying mathematical and physical modeling and applying three essential machine learning baselines (Logistic Regression, Neural Networks and Gradient Boosting on Decision Trees) for the problem of lithotype classification based on MWD data, which come from 27 wells of the Novoportovskoe oil and gas condensate field on Western Siberia.

\subsection{Machine Learning in drilling application} 

There are previous studies on the involvement of machine learning for detection of a material type at drilling bit. \cite{zhou2010hybrid} cover an analysis of the applicability of regression and classification based on Gaussian Processes and unsupervised clustering for on-bit rock typing with MWD data. In the report the authors consider the rate of penetration (\ROP{}), pulldown pressure, which is also referred after as a weight on bit (\WOB{}), and top drive torque (\TRQ{}) as the key parameters for building the data-driven forecasting model. One of the conclusions is that a value called adjusted penetration rate (\APR{}) (Eq. \ref{eq:APR}) is the best reflection of a features specifics of the rock which are unknown a-priori. The authors conclude that the optimal way to predict a rock type at the drilling bit is to apply a hybrid model combining the advances of both supervised classification and unsupervised clustering. 

\begin{equation} \label{eq:APR}
	\begin{split}
		{\APR{} \propto \cfrac{\ROP{}}{{\WOB{}}\sqrt{\TRQ{}}}}
	\end{split}
\end{equation}

\APR{} is tested in this study as well as another characteristic utilized by many authors \citep{zhou2010hybrid, zhou2011adaptive}, the Specific Energy of Drilling (\SED{}):

\begin{equation} \label{eq:SED}
	\begin{split}
		\SED{} = \frac{\WOB{}}{A} + \frac{2\pi \times \RPM{} \times \TRQ{}}{A \times \ROP{}},
	\end{split}
\end{equation}
where $A$ represents a cross section area of the wellbore. 

\cite{zhou2011adaptive} illustrates that unsupervised learning together with the minimization of \SED{} is a promising approach for the optimization of the penetration rate. Another effort on penetration rate optimization is presented by \cite{hegde2017use}. The authors use the Random Forest algorithm to build a model linking the penetration rate with weight on bit, rotation speed, drilling mud rate, and unconfined rock strength. The model allowed to optimize the penetration rate for up to $12\%$ for the wells close to ones in the training set. 

\cite{labelle2000material} and \cite{labelle2001lithological} describe an application of Artificial Neural Networks for material typing and rock typing at drilling. MWD-like measurement and the trained Neural Networks allow a relative classification error to be as small as $4.5\%$ for a case with the complete set of available mechanical measurements (features). 

According to the fundamentals of Machine Learning, Gaussian Processes and Neural Networks are not the best fit for rock type classification with MWD data as they can not automatically handle missing values that typically occur in MWD data. Thus, both methods require training data without missing values that implies the development of accurate imputation procedures. The difference between these methods is in the preferred data size and its dimensionality. Gaussian Processes are based on the Bayesian approach, so they can work well when training sample is small, however, their area of application is limited to low input dimensions and small sample sizes (up to 10000 elements). Neural networks are based on frequentist inference, so they require large training samples, but they can work well in large dimensions. In case we need to reflect the temporal behavior of MWD in input features, we face high dimensions, also for real-life MWD sample sizes are large. Therefore Neural Networks would be more preferable than Gaussian Processes, if there are no missing values in training and real-life data.

Decision trees and methods based on them \citep{hastie_09_elements-of.statistical-learning} such as Random Forest and Gradient Boosting can automatically handle missing values, and they are comfortable with large sample sizes. However, tree-based methods are weak at data interpolation, so they generalize well only when the density and the diversity of points in the training sample are high. Gradient Boosting can also handle classes imbalance by automatic weighting the importance of data entries while maximizing the quality of a classifier.

\subsection{Modeling of Drilling Mechanics}

Physical models are based on the physical equations (typically mass and energy balances) governing the system under analysis. 
\cite{Downton2012} examines the modeling of different aspects of drilling and focuses on the possibility of bringing these models together into a single approach and creating unified control systems to automate the entire process.
\cite{Sugiura2015} gives 
the most accurate description of the state-of-the-art 
in the modeling of drilling systems for automation and control, adaptive modeling for downhole drilling systems and actual tasks of the industry.
\cite{Cayeux2014} provides a detailed analysis
of sensor equipment on the drilling rig and the issues of its layout based on obtaining the most qualitative boundary and initial conditions for solving the problems of physical modeling of the drilling process.
The majority of the papers on drilling mechanics are devoted to the vibrational analysis of the drill-string \citep{Shor2014}.

Initially, analytical formulas derived from a simplified view of the drilling process can be used \citep{detournay1992}. The input data (\WOB{} and \RPM{}) allow to predict the output (\TRQ{} and \ROP{}). The main difficulty is the calibration of the model, which requires finding the model coefficients from the experimental data. The general scheme is the following. For a known set of lithotypes in height with unknown parameters of the model, a numerical solution is found, and the computed values of \ROP{} are compared with the experimental data.
Thus, in the presence of a sufficient number of experimental data, it is possible to find the model coefficients for each of the lithotypes and bit types. 
Therefore, one may simulate the drilling process for an arbitrary set of lithotypes in height, thereby substantially expanding the training set for the predictive model.

Non-linear models of drill string vibrations were considered by \cite{Spanos2002}, where the nonlinearity 
arises when taking into account the interaction of the bit and rock. Only lateral vibrations were examined therein.
The state of the system is described by the transverse displacement $u$ and the angle of rotation $\theta$ of each of $N$ segments. The resulting system of equations is: 
\begin{equation}\label{mur:spanos:maineq}
M \mathbf{u}'' + C \mathbf{u}' + K \mathbf{u} + F(\mathbf{u}) = g(t), 
\end{equation}
where $\mathbf{u}=[u_1,\dots,u_N, \theta_1,\dots,\theta_N]$; $M$, $C$, $K$ are the system mass, damping and stiffness matrices, respectively; $g(t)$ denotes the excitation applied to the system, and $\mathbf{u}, \mathbf{u}', \mathbf{u}''$ correspond to the displacement, velocity and acceleration vectors.
Nonlinear part $F(u)$ plays an important role, it arises 
due to the contact interaction of the drill string with the wall. While matrices $M,C,K$ depend on properties of drill-string, the friction term $F$ depends on rock type. By solving the inverse problem for $F$, for example, determining constants in Hertzian contact law, we get parameters characteristic for rock type. To increase the quality of the model the right-hand side of equation \eqref{mur:spanos:maineq} can be considered as a random (Wiener) process. 
Unfortunately, this type of models is hardly applicable 
as input data is incomplete: to get matrices $M,C,K$ we need to know exact geometric properties of drill-string along with material ones. 

\section{Materials and Methods}
This section first specifies the origin of data used in our work and its pre-processing procedures, next it describes machine learning methods we studied for rock type classification at a drilling bit, then the section defines quality metrics used for evaluation of classifiers, and finally, it describes approaches for improving classification quality by choosing input features. 

\subsection{Data description and pre-processing}
This subsection specifies geological formation on which the data was collected, then it outlines essential for this work components of the data, and describes the process of obtaining them from the raw exported files. 

\subsubsection{Geological formation of the interest}
The Novoportovskoye oil and gas condensate field, located within the Yamal Peninsula, 30 km from the Gulf of Ob Bay, is the largest field under the development of the northwest of Siberia, Russia. 
The formation includes several strata, the most productive of which is the Lower Cretaceous NP-2-3 --- NP-8 (the formation depth is 1800 m), and sand layers of the Tyumen suite J-2-6 (the formation depth is 2000 m). The reservoir rocks are fine-medium grained sandstones and siltstone with thin layers of shales and limestone. The average rocks permeability is 0.01-0.03 $\mu$m$^2$ and the porosity is about $18\%$. 

\subsubsection{Initial data}

The initial data included mud logging, involved the rig-site monitoring and assessment of information measured on the surface while drilling and MWD, LWD data from downhole sensors.
The main purpose of MWD systems is to determine and transmit to the surface of the inclinometry data (zenith angle and magnetic azimuth) in real time while drilling. It is necessary to determine the well trajectory. Sometimes the inclinometry data are supplemented with information about the drilling process and logging data (LWD). Logging allows measure the properties of the rock, dividing the geological section into different lithotypes.

The data includes the following parameters: \WOB{}, \TRQ{}, \ROP{}, \APR{}, \SED{}, also rotary speed (\RPM{}), input flow rate (\Qin{}), output flow rate (\Qout{}), standpipe pressure (\SPP{}), and hook load (\HL{}). 

Initial information about the drilled lithotypes was Lithology Map produced by petrophysical interpretation of LWD measurements which were represented by natural gamma radiation; apparent resistivity; polarization resistance; electromagnetic well log; induced gamma-ray log; neutron log; acoustic log.

LWD petrophysical interpretation was also used to compare the real values of the lithotype and the prediction obtained.

\subsubsection{Pre-processing}
\label{sss:preprocessing}

For the solutions based on machine learning approaches, it is crucial to preprocess raw data into a suitable format for data-driven algorithms, also known as constructing data-pipeline. For the real-world cases, the problem of preprocessing is usually complicated: the size of the raw data, the variety of formats and the number of sources can be too large to apply straightforward methods \citep{Garcia2016, 7207219}. Although some formats are common for oil-and-gas industry, such as .las files, others can vary from company to company e.g. drilling reports. 
 In order to effectively process source files, the pipeline has to handle common types of errors in them. Some formats can also have different options, for example, different .csv files can have different columns separators.
Moreover, the number of wells for the preprocessing can be as large as hundreds or even thousands, so the proposed framework should work in a fast and accurate way in an automatic regime.

In this study we used a task-based approach using Luigi\footnote{\texttt{https://pypi.python.org/pypi/luigi}} framework for Python programming language. This framework allows building data pipelines, where each step of the preprocessing can be implemented as a separate task, such as processing of source files or merging some chunks of data, which can depend on other tasks. Thus, the whole pipeline is resistant to errors in raw data and dependencies between tasks. 

\begin{figure}
\centering
\includegraphics[width=0.99\textwidth]{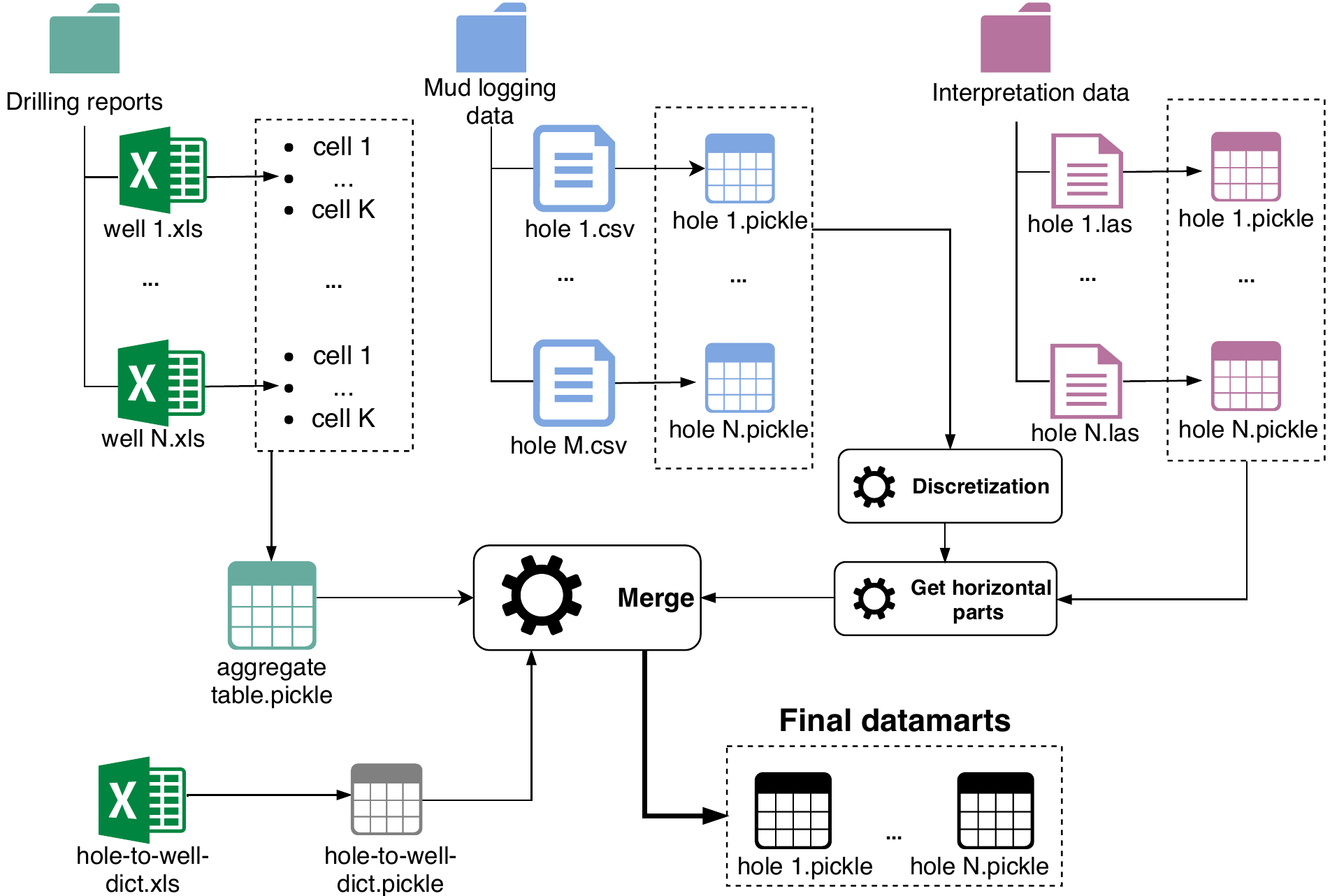}
\caption{Raw data preprocessing pipeline.}
\label{fig:preprocessing_schema}
\end{figure}

\paragraph{Pipeline description}
The complete scheme of pipeline used in our study is shown in Figure \ref{fig:preprocessing_schema}. 
The pipeline for the preprocessing of the data consisted of four main steps:
extraction of required columns from the raw data files; selection of the relevant horizontal parts of the wells; merging data from different sources; unifying depths steps for constructed dataframes. 

All data sources were in different directories. 
The first step for each source file processing was the extraction of required columns or cells of data. 
This step is represented in the schema as outgoing arrows from each file (\texttt{.xls}, \texttt{.las} or \texttt{.csv}). 
All information from drilling reports was aggregated into the file \texttt{aggregate\_table.pickle}. 
We stored the results of each intermediate step in pickle files, which were serialized tables of data. 
Pickle format is storage-consuming, but fast for input/output operations. 

The mud logging data was discretized by files with the sampling frequency equal to the sampling of other sources of data in block "Discretization". Next, we extracted data corresponding to the horizontal part from each mud logging table in the block "Get horizontal part". 
For obtaining boundaries of the horizontal parts, we used the interpretation data. 

Some wells had several laterals (in preprocessing pipeline they were called holes), that is why part of data was associated with laterals (e.g. mud logging data), and other data was connected to wells (e.g. drilling reports).
The final step of the preprocessing is merging data for each hole by depth (see block "Merge" in the schema). 
For merging all chunks of data, we used a table with the correspondence between laterals and wells from \texttt{"hole-to-well-dict.xls"}. 
As a result, we received the set of merged data into depth-associated time series by laterals (see block "Final datamarts" in the schema). 

After preprocessing of the raw data, we reduced the granularity of time-series by aggregating them over depth intervals of size 0.1 meters. For intervals containing any data, we averaged its values, for intervals without data, we used forward fill with a constant that equals the latest preceding value.

\subsection{Machine Learning Models}

We considered the of rock type identification as the common machine learning binary classification problem. To attack this problem, we used three machine learning methods: Logistic Regression, Gradient Boosting on decision trees and Artificial Neural Networks. These methods are described in this section.

\subsubsection{Logistic regression}

The logistic regression is a generalization of the linear regression to classification problem~\cite{hastie_09_elements-of.statistical-learning}.

For logistic regression, we suppose that the target probability of an object to belong to a certain class is a sigmoid transformation $\sigma(\eta) = 1 / (1 + \exp(-\eta))$ of a linear function of input features $\eta = \sum_{i = 1}^d w_i x_i$, where $x_i$ is a value of some input feature e.g. WOB, and $w_i$ is a weight for this feature.
During the learning phase, we estimate weights $w_i$ by maximizing likelihood or quality of fit of the model to the data. 

In this article we use Logistic regression as a baseline to identify improvement due to the usage of more complex significantly nonlinear Gradient Boosting and Artificial Neural Networks approaches for our problem.

\subsubsection{Decision trees and Gradient Boosting}

\begin{figure}
\centering
\includegraphics[width=0.99\textwidth]{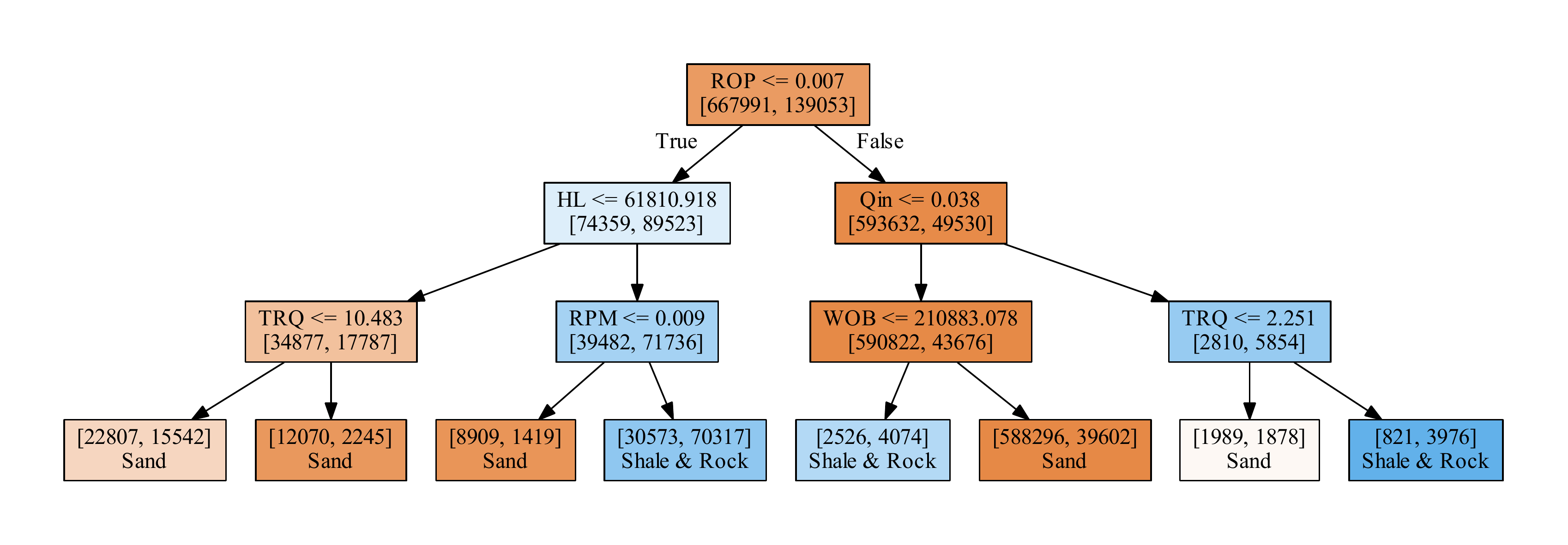}
\caption{An example of a real decision tree for the lithotype classification: internal nodes contain decision rules, the splits of the training objects that fall into this node into two classes (Sand --- left number, Shale \& Rock - right number).
Color of the node corresponds to this distribution. Leaf nodes don't have decision rules, but provide suggested classes.} 
\label{fig:example_decision_tree}
\end{figure}

The most widely used approach for the solution of classification problems is based on the Ensembles of Decision Trees. 
An example of a decision tree is presented in Figure~\ref{fig:example_decision_tree}:
\begin{itemize}
\item for each object the classifier proceeds through the decision tree according to the values of input variables for this object until it reaches a leaf of the tree, 
\item if it reaches the leaf, it returns either the major class in this leaf or probabilities to belong to classes according to the distribution of objects of different classes from the training sample, that correspond to this leaf.
\end{itemize}
The advantages of this approach include a superior performance with default settings~\citep{fernandez2014we}, fast model construction, almost no over-fitting and handling of various problems in data including the availability of missing values and outliers.

Among various approaches for construction of Ensembles of Decision Trees, the most popular nowadays is Gradient Boosting~\citep{chen2016xgboost,kozlovskaya2017deepboost}, which essentially follows functional gradient in the space of decision tree classifiers to construct the ensemble. At each step it increases weights of objects that are poorly classified using the current ensemble, thus increasing their contribution to the total model quality measure. The algorithm has the following main parameters: 
\begin{itemize}
\item learning rate --- how fast it learns the ensemble. If learning rate is too small, we need to use larger number of trees in the ensemble at the cost of larger computational power, which grows linearly from the number of trees; in the opposite case, we can get overfitting as the adaptation of the ensemble to the training data occurs too fast; In the experimental section we demonstrate this effect in Figure \ref{fig:ntrees_vs_lr};
\item maximal depth --- maximal depth for each tree in the ensemble;
\item random subspace share --- share of features used by each decision tree;
\item subsample rate --- share of objects from the training sample used for training of each decision tree.
\end{itemize}

\subsubsection{Artificial Neural Networks}

Alternative modern data-driven approach for classification problems is Artificial Neural Networks.
They are more demanding for quality and size of input data and require more subtle tuning of hyperparameters.
On the other hand, this type of machine learning algorithms can be more powerful in some types of problems and for some specific structures of input data \citep{8037515, AHMAD201777}.

The main idea behind Neural Networks is to define a deep composition of sequential application of linear and nonlinear multi-input and multi-output functions parameterized by weights of linear functions. 
Each composition of linearity and nonlinearity is called a layer.
As gradient of classification error is easy to propagate through this composition, we can apply gradient methods for optimizing a quality metric with respect to these parameters and get a pretty accurate model in the end.

\begin{figure}
\centering
\includegraphics[width=0.7\textwidth]{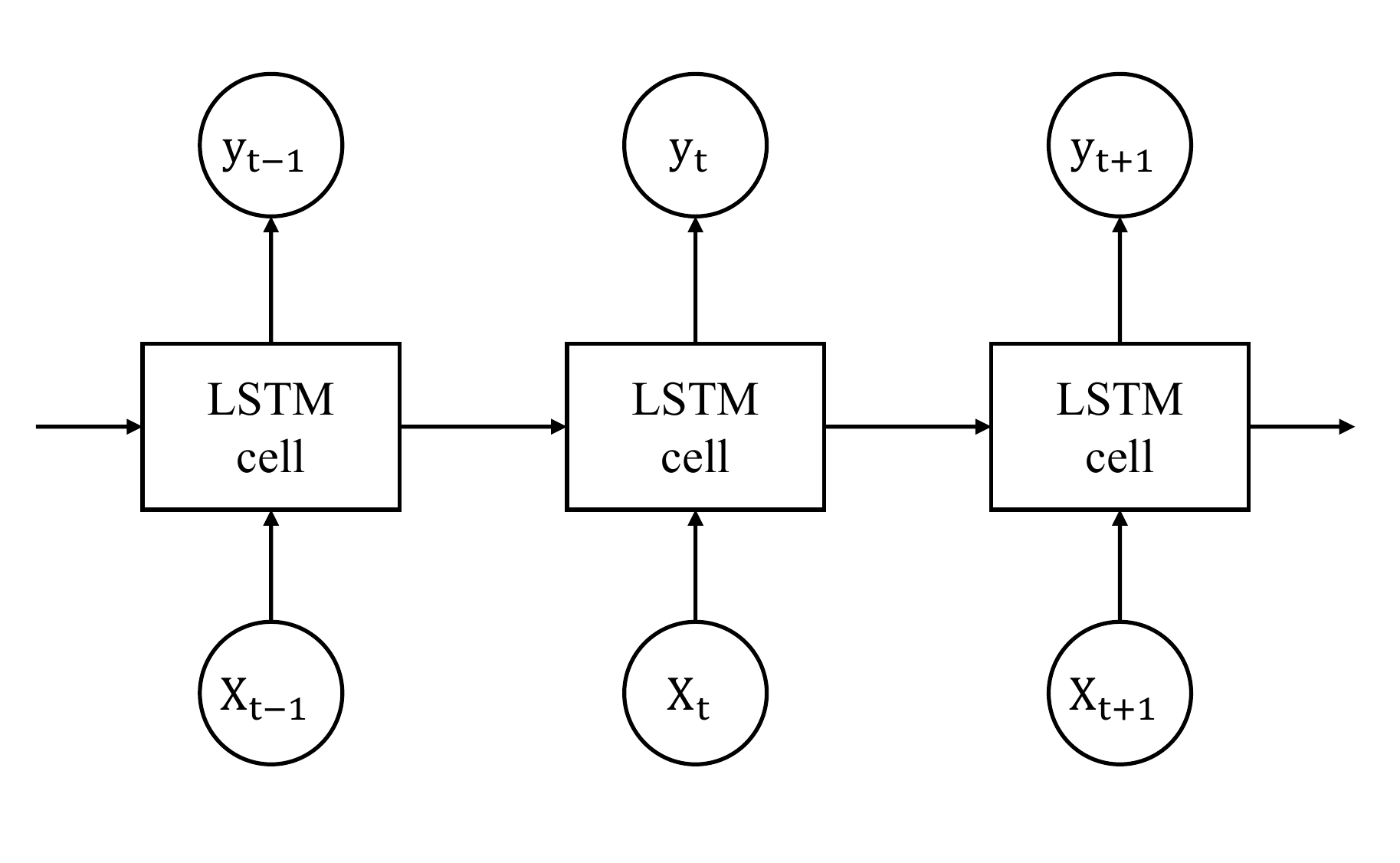}
\caption{An illustration of information flows in LSTM. $X_t$ are input values at the moment $t$, $Y_t$ is the corresponding output of the network, arrows between LSTM cells represent additional input of internal variables from the previous moment.} 
\label{fig:example_lstm}
\end{figure}

There are many ways to define this deep composition, the most relevant to our problem are Feed Forward fully connected~\citep{Hornik:1989:MFN:70405.70408} and Long-Short Term Memory (LSTM) ~\citep{Hochreiter:1997:LSM:1246443.1246450} architectures.
For fully connected architecture we connect each input with each output at each layer, when applying linear function.
For LSTM we use as additional input some variables from the previous moment of time, thus keeping some information from the distant pass and creating long-term memory effect for a neural network. This scheme is shown in Figure \ref{fig:example_lstm}.

During experiments we applied two classes of Neural Networks: Feed Forward and LSTM.
Our experiments were based on different configurations of these classes of Neural Networks.

\subsection{Quality metrics}

There are many quality metrics for comparing classifiers.
In this article we used three metrics: 
a specific industry-driven metric Accuracy L and 
two common machine learning metrics, namely, area under Receiver Operating Characteristic (ROC) curve (ROC AUC) and area under Precision-Recall (PR) curve (PR AUC).

We used additional quality metrics, because accuracy metric alone is not very representative due to significant class imbalance, such that a constant "always-sand" predictor gives relatively high accuracy, yet brings no practical benefits. 

We did not consider specific metrics for time-series or ordered data, 
as there was no universally acknowledged metric that was easy to interpret~\citep{burnaev2017automatic,artemov2015ensembles}.

Let us consider a test sample $D = \{(\mathbf{x}_i, y_i)\}_{i = 1}^n$, $\mathbf{x}_i$ is an input vector for an interval, $y_i$ is a true class, either $0$ (Sand) or $1$ (Shale and Rock).
We have predictions by a classifier for each interval $\hat{y}_i \in \{0, 1\}$.
The length of each interval is $l_i$, $i = \overline{1, n}$.

Accuracy L is the sum of lengths of intervals with correct predictions of lithotype divided by the total depth of considered wells.
\begin{equation}
\mathrm{Accuracy\, L} = \frac{\sum_{i = 1}^{n} l_i [y_i = \hat{y}_i]}{\sum_{i = 1}^{n} l_i},
\end{equation}
where for any arguments $a$ and $b$ expression $[a = b]$ henceforth means the indicator function:
it equals $1$ if $a$ is equal to $b$, and $0$ otherwise.

To define ROC AUC and PR AUC metrics we need to introduce additional notation. After training a classifier, it outputs a probability of an object to belong to a class.
To obtain the final classification with labels we apply a threshold to the probabilities: the objects with probabilities below the threshold are classified as the first class objects, and the objects with probabilities above the threshold are classified as the second class objects.

For a particular classification there are four numbers that represent its quality: number of True Positive (TP) --- correctly classified objects of the first class, False Negative (FN) --- objects of the first class attributed by the classification to the second class , False Positive (FP) --- -- objects of the second class attributed by the classification to the first class, and True Negative (TN) --- correctly classified objects of the second class:
\begin{align}
\mathrm{TP} &= \frac{1}{n} \sum_{i = 1}^n [y_i = 1] [\hat{y}_i = 1],
\mathrm{TN} = \frac{1}{n} \sum_{i = 1}^n [y_i = 0] [\hat{y}_i = 0], \\
\mathrm{FP} &= \frac{1}{n} \sum_{i = 1}^n [y_i = 0] [\hat{y}_i = 1],
\mathrm{FN} = \frac{1}{n} \sum_{i = 1}^n [y_i = 1] [\hat{y}_i = 0].
\end{align}
By dividing the number of TP objects by the total number of positive objects (sum of TP and FN) we get True Positive Rate (TPR),
by dividing the number of False Positive objects by the total number of negative objects (sum of False Positive and True Negative objects) we get False Positive Rate (FPR):
\begin{equation}
\mathrm{TPR} = \frac{\mathrm{TP}}{\mathrm{TP} + \mathrm{FN}}, \mathrm{FPR} = \frac{\mathrm{FP}}{\mathrm{FP} + \mathrm{TN}}.
\end{equation}

By varying the threshold, we get a trajectory in the space of 
TPR and FPR that starts at point $(0, 0)$ when all objects are classified to the negative class, and ends at $(1, 1)$ where all objects are classified to the positive class.
This trajectory is ROC curve. 
In a similar way we define precision as $\mathrm{TP} / (\mathrm{TP} + \mathrm{FP})$ and recall as $\mathrm{TP} / (\mathrm{TP} + \mathrm{FN})$ and plot the trajectory in the space of precision and recall.
This trajectory is PR curve.

By calculating areas under ROC and PR curves, we get correspondingly ROC AUC and PR AUC widely used to measure the quality of classifiers.
Higher values of ROC AUC and PR AUC suggest that the classifier is better. ROC AUC and PR AUC values for a random classifier are $0.5$ and the share of the positive class respectively, ROC AUC and PR AUC values for the perfect classifier are $1$.
For imbalanced classification problems, PR AUC suits better,
for a detailed discussion on metrics for imbalanced classification see \cite{burnaev2015influence} and references therein.

\subsection{Feature engineering and selection}
\label{sec:feature_engineering_and_selection}

In this section we describe several methods of refining information about rock types which is stored in MWD and LWD data, so that classifiers can take advantage of it.
 
\subsubsection{Time-series features}
At each moment of time not only current MWD and LWD values characterize the type of rocks, but also previous values and their relationships among each other bring additional information. Therefore in this section, we start with considering a few ways to incorporate such information as input features.

The \emph{Basic} (B) set of features used in a predictive model includes original mechanical features, \SED{}, and \APR{}. We also derived new features from the basic ones:
\begin{itemize}
\item \emph{Derivatives} (D) --- rolling mean and standard deviation with the window size of 1 m, and the difference between values on rolling window's borders;
\item \emph{Lagged} (L) --- lagged basic features i.e. their values 0.1, 0.5, 1 and 10 m ago;
\item \emph{Fluctuations} (F) --- standard deviation of original time series inside aggregated (see sec. \ref{sss:preprocessing}) intervals of 0.1 m;
\item \emph{Extra} (E) features --- true class values 20 and 50 m ago, since they can be obtained from LWD measurements with such spatial lags.
\end{itemize}

\subsubsection{Mathematical modeling of drilling mechanics}
Rock destruction under load has been studied in great detail by \cite{mishnaevsky1993} and \cite{mishnaevsky1995}, but only a few works studied dynamic properties of the process.

We started with the assumption that 
drill-bit rock interaction could be described as several processes:
rock crushing, rock cutting and rotary friction on drill-bit. 
We further assumed the rate of penetration was proportional to the weight on bit (rock crushing part) 
and the angular velocity $\Omega$ (cutting and friction part):
\begin{equation}
\label{rop}
\mbox{\ROP{}} = a_1 + a_2 \mbox{\WOB{}} + a_3 \Omega.
\end{equation}
On the other hand, following \cite{detournay1992} 
and assuming torque on bit is mainly related to rock cutting process, we had the following relation:
\begin{equation}
\label{tob}
\mbox{TOB} = a_4 \frac{\mbox{\ROP{}}}{\Omega} + a_5.
\end{equation}

To get a smaller set of parameters, we substituted (\ref{rop}) into (\ref{tob}):
\begin{equation}
\label{model}
\mbox{TOB} = \frac{b_1 + b_2 \mbox{\WOB{}}}{\Omega} + b_3.
\end{equation}
For the fixed bit, parameters $b_1, b_2, b_3$ depend on rock properties and therefore can characterize them, so they can be used as \emph{Mathematical} (M) features for rock type identification. 
These parameters were obtained for short intervals with size $m$ of MWD time-series by solving the optimization problem \eqref{eq:math_opt}, which minimized the model local discrepancy at some moment $k$:
\begin{equation}
\label{eq:math_opt}
b_1(k), b_2(k), b_3(k) = \underset{b_1, b_2, b_3}{\mbox{argmin}} \sum_{i = k - m + 1}^{k}  \left(\mbox{TOB}_k - \frac{b_1 + b_2 
\mbox{\WOB{}}_k}{\Omega_k} - b_3 \right)^2.
\end{equation}
Because of locality, window size $m$ should not be large.

\subsubsection{Feature selection}

Generating too many interrelated features results in their redundancy, longer time of models training and risk of overfitting. Thus, after feature engineering, we ran the feature selection procedure which had the aim to select the subset of features that maximized classification quality. 

We used a "greedy" approach for feature selection: the procedure started from the empty set and expanded it by adding step by step the most impactful feature from the pool of remaining ones according to a selected quality metric.

\section{Results}

In this section we:
\begin{itemize}
    \item report on how different sets of features affect the quality of rock type classification, which features are more informative;
    \item examine selection of hyperparameters for different machine learning methods;
    \item compare the performance of different machine learning methods and show how classification quality depends on the balance of classes.
\end{itemize}. 

\subsection{Feature selection results}
\label{sec:feature_selection}

For feature selection we used ROC AUC quality metric obtained via leave-one-well-out cross-validation (LOWO-CV). Since sensors readings are autocorrelated, it is crucial to split the dataset by wells, not by random subsets during cross-validation. Otherwise, data leakage will take place resulting in overestimated quality, that is, models will have more information about the test set during cross-validation than they will have in the field test on new wells.

The classifier was constructed with Gradient Boosting of $50$ decision trees, each of maximal depth $6$. 
The best selected set \emph{Greedy} (G) consists of \ROP{}, \HL{}, rolling differences of \WOB{}, 1m rolling standard deviations of \ROP{} and \TRQ{}, 1m moving average of \ROP{}, $0.5$ meters lagged \TRQ{}, and $10$ meters lagged \Qout{}, \Qin{}, \HL{} and \TRQ{}. 

We also fine-tuned Gradient Boosting hyperparameters by increasing the number of decision trees up to 100 and conducting a grid-search LOWO-CV for maximal depth of trees, random subspace share and sub-sampling rate.
Table \ref{tb:feature_selection} summarizes the results of the feature selection process. 
We obtained the best results for all quality metrics using the selected set of features G along with extra set E. In particular, Accuracy L is larger than $0.9$.


\begin{table}[H]
\setlength{\tabcolsep}{2pt}
\centering
\begin{tabular}{llll}
\hline
Feature set & ROC AUC & PR AUC & Accuracy L \\
\hline
- & 0.494 & 0.181 & 0.866 \\
B &  0.794 & 0.492 & 0.865 \\
B, F &  0.803 & 0.484 & 0.867 \\
B, F, D, L &  0.829 & 0.504 & 0.870 \\
G &  0.850 & 0.559 & 0.888 \\
E & 0.653 & 0.359 & 0.879 \\
B, E &  0.848 & 0.581 & 0.900 \\
B, F, D, L, E &  0.870 & 0.600 & 0.902 \\
G, E &  $\mathbf{0.878}$ & $\mathbf{0.614}$ & $\mathbf{0.905}$ \\
\hline
G, E (fine-tuned) & $\mathbf{0.880}$ & $\mathbf{0.625}$ & $\mathbf{0.910}$\\
\hline
\end{tabular}
\caption{Feature selection results. Greedy selected set of features combined with the Extra set provides the best quality.}
\label{tb:feature_selection}
\end{table}

\begin{figure}[H]
\centering
\begin{subfigure}[b]{.48\linewidth}
\includegraphics[width=1\textwidth]{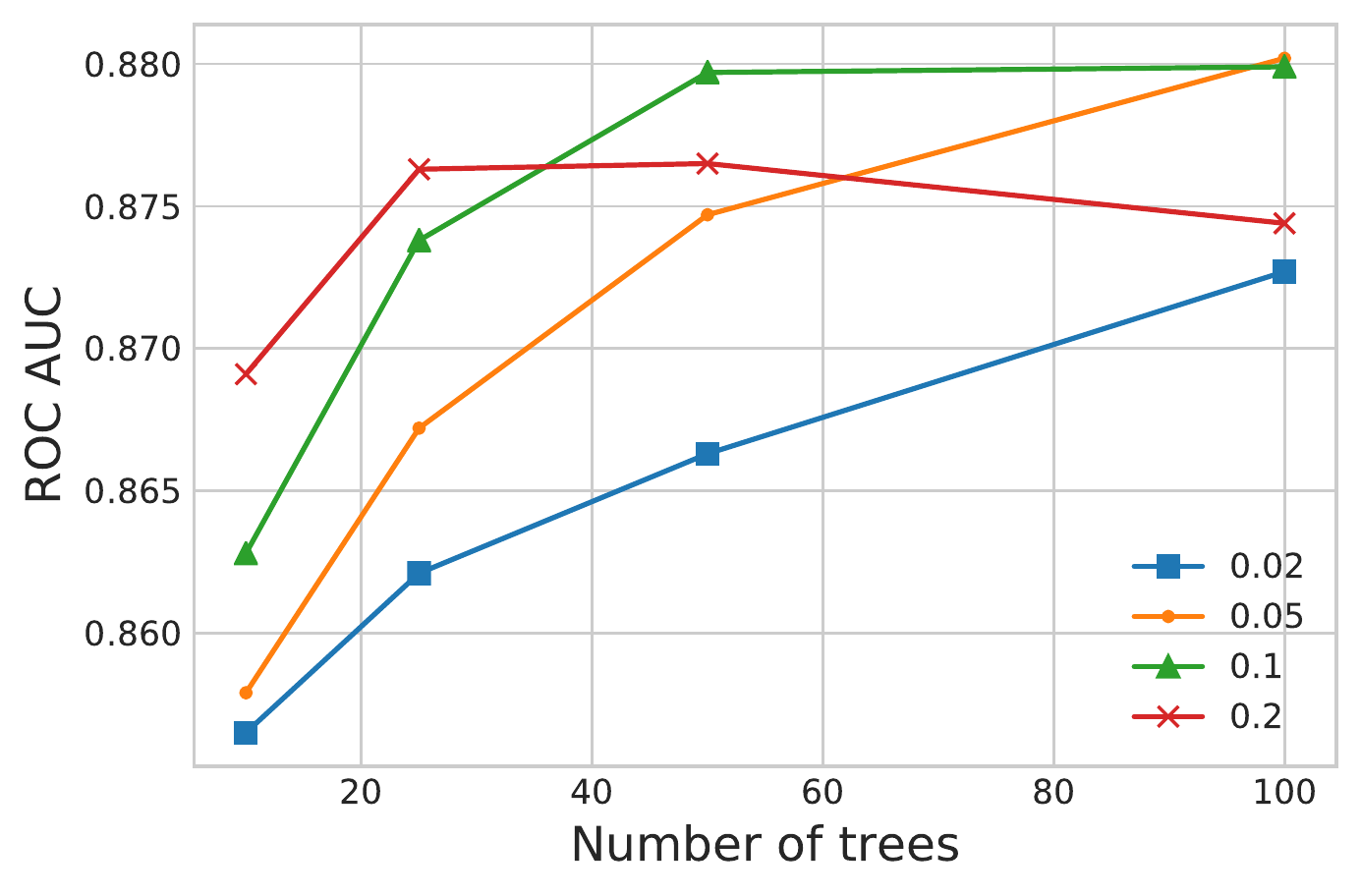}

\end{subfigure}
\begin{subfigure}[b]{.48\linewidth}
\includegraphics[width=1\textwidth]{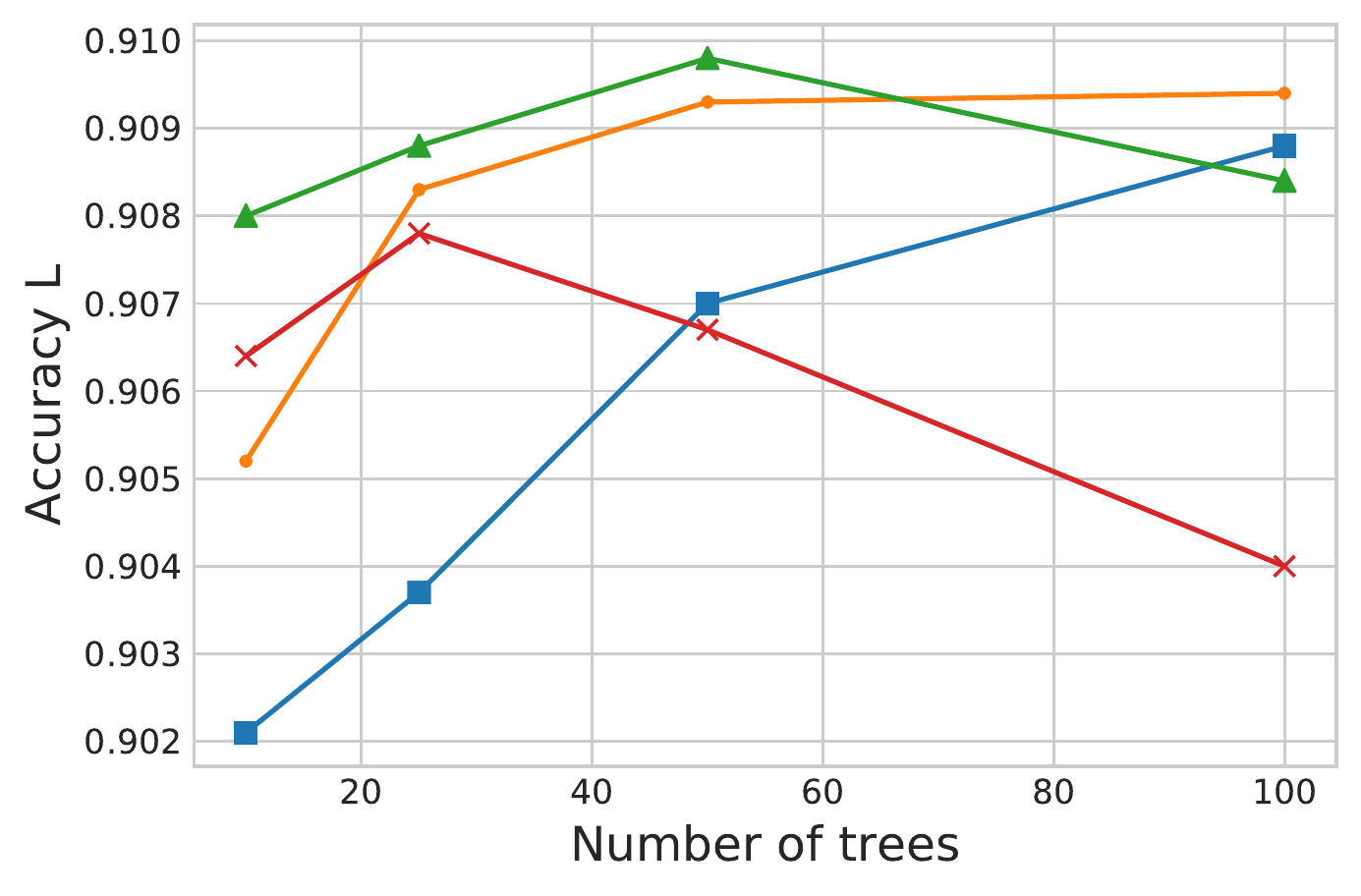}

\end{subfigure}
\caption{Quality vs Gradient Boosting parameters. Curves of different colors correspond to different learning rates.}
\label{fig:ntrees_vs_lr}
\end{figure}

Figure \ref{fig:ntrees_vs_lr} shows the dependence of quality metrics on learning rate and the number of trees in the ensemble obtained by Gradient Boosting. Low learning rates (blue curves) result in underfitting, whereas high learning rates (red curves) result in overfitting of the model. Orange and green curves correspond to a good trade-off.

\begin{figure}[H]
\centering
\includegraphics[width=0.9\textwidth]{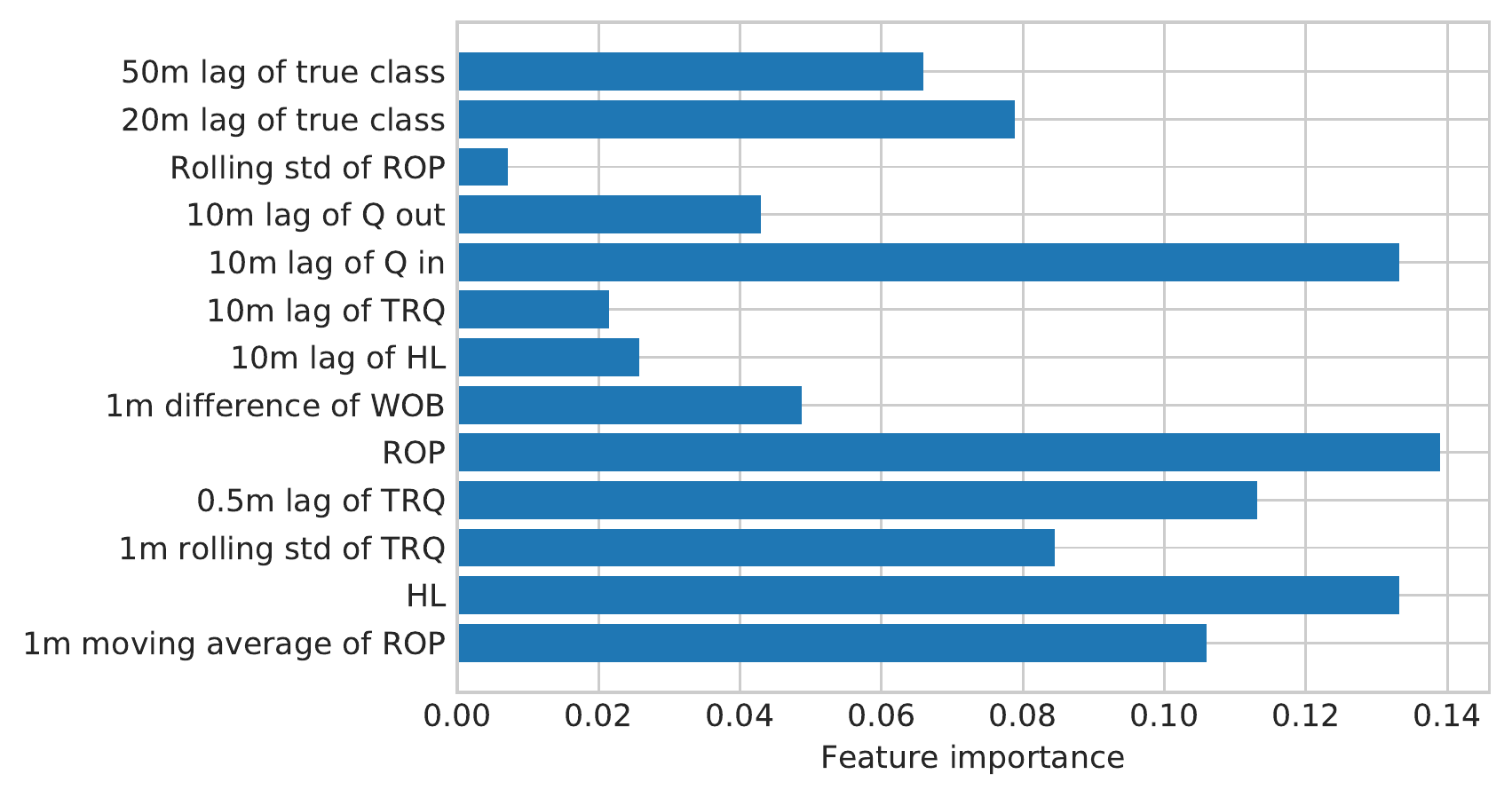}
\caption{Importance of features for the Gradient Boosting classifier predictions. Two sets of features are included: Greedy and Extra. The bottom-up order of Greedy features corresponds to their selection order during the selection procedure.}
\label{fig:greedy_feature_importance}
\end{figure}

Figure \ref{fig:greedy_feature_importance} shows feature importances for the fine-tuned classifier trained on the whole dataset. Importance scores indicate how many times a particular feature played the key role in the classifier's decision.

\subsection{Examination of mathematical modeling features}

Only 13 out of 27 wells had no missing values of features required for mathematical modeling.
For them we studied the effect of Mathematical features (M) and their fluctuations (FM) on quality metrics. We used window size $m=5$. The results are presented in Table \ref{tb:feature_phys}. 
Mathematical modeling features turned out to have weak predictive power: no significant gain on top of the Greedy features was obtained.


\begin{table}[H]
\setlength{\tabcolsep}{2pt}
\centering
\begin{tabular}{llll}
\hline
Feature set & ROC AUC & PR AUC & Accuracy L \\
\hline
- & 0.499 & 0.198 & 0.858 \\ 
B &  0.837 & 0.552 & 0.890 \\
M &  0.524 & 0.208 & 0.829 \\
M, FM &  0.566 & 0.264 & 0.855 \\
G &  0.874 & $\mathbf{0.609}$ & $\mathbf{0.906}$ \\
G, M &  $\mathbf{0.875}$ & 0.597 & 0.904 \\
G, M, FM &  0.870 & 0.590 & 0.900 \\
\hline
\end{tabular}
\caption{Performance of the approaches significantly depends on the set of used features. However, the usage of mathematically modeled features doesn't improve quality.}
\label{tb:feature_phys}
\end{table}

\subsection{Algorithms performance}

We compared three classes of machine learning methods in details: Logistic regression, Gradient Boosting, and Neural Networks. Results in this section correspond to the performance of the best-found configurations for each method using LOWO-CV. All methods used both Greedy and Extra sets of features.

For logistic regression, we observed the best quality when no regularization is applied. 
The best-found configuration of Gradient Boosting for 100 trees had the following hyper-parameters: learning rate 0.05, maximal depth 3, random subspace share 0.8, and sub-sampling rate 0.55. 
For Feedforward Neural Networks (NN) we tested different architectures with 2-, 3- and 4-layer networks. The best found configuration had two hidden layers of size 100 and 500 neurons using ReLU activation between layers. 

Table \ref{tb:algorithms_comparison} summarizes the best performance of different classification methods. Gradient Boosting uniformly dominates logistic regression, in turn, Feedforward NN and Gradient Boosting qualities are comparable due to the preprocessing pipeline we developed, which filled the missing data sections with rather adequate values. LSTM training time was impractically long, whereas its best-found performance was similar to Feedforward NN.


\begin{table}[H]
\setlength{\tabcolsep}{2pt}
\centering
\begin{tabular}{llll}
\hline
Algorithm & ROC AUC & PR AUC & Accuracy L \\
\hline
Always predict the major class & 0.494 & 0.181 & 0.866 \\
Logistic regression & 0.860 & 0.585 & 0.908  \\
Gradient Boosting & $\mathbf{0.880}$ & $\mathbf{0.625}$ & 0.910\\
Feedforward NN & 0.875 & $\mathbf{0.625}$ & $\mathbf{0.911}$ \\ \hline
\end{tabular}
\caption{Performance of machine learning approaches Logistic regression, Gradient Boosting, and Feedforward NN. All performance measures are better if higher.}
\label{tb:algorithms_comparison}
\end{table}

\begin{figure}[H]
\centering
	\begin{subfigure}[b]{.49\linewidth}
      \includegraphics[width=1.0\textwidth]{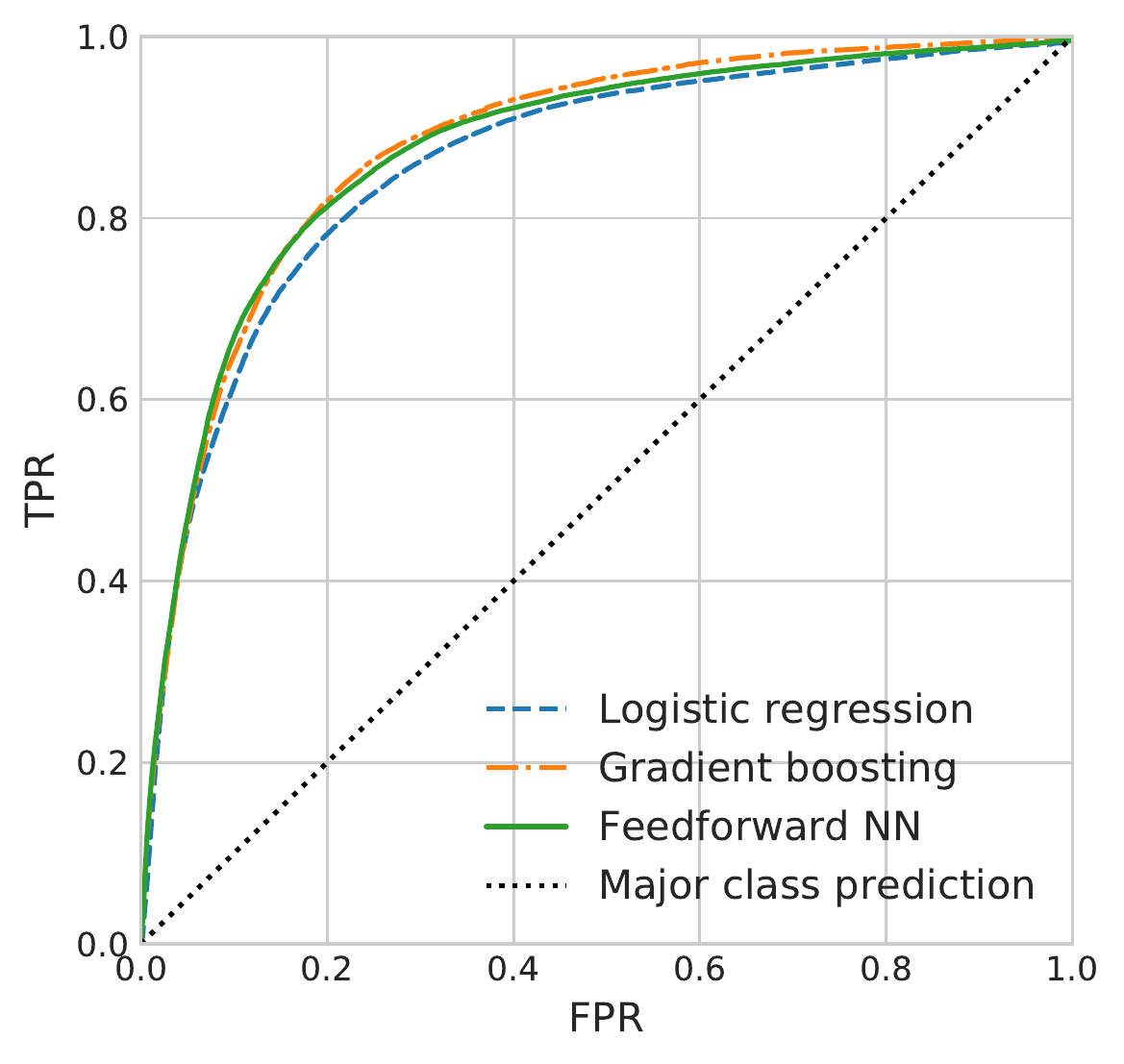}
      \caption{ROC curve}
      \label{fig:roc_curve}
	\end{subfigure}
    \begin{subfigure}[b]{.49\linewidth}
      \includegraphics[width=1.0\textwidth]{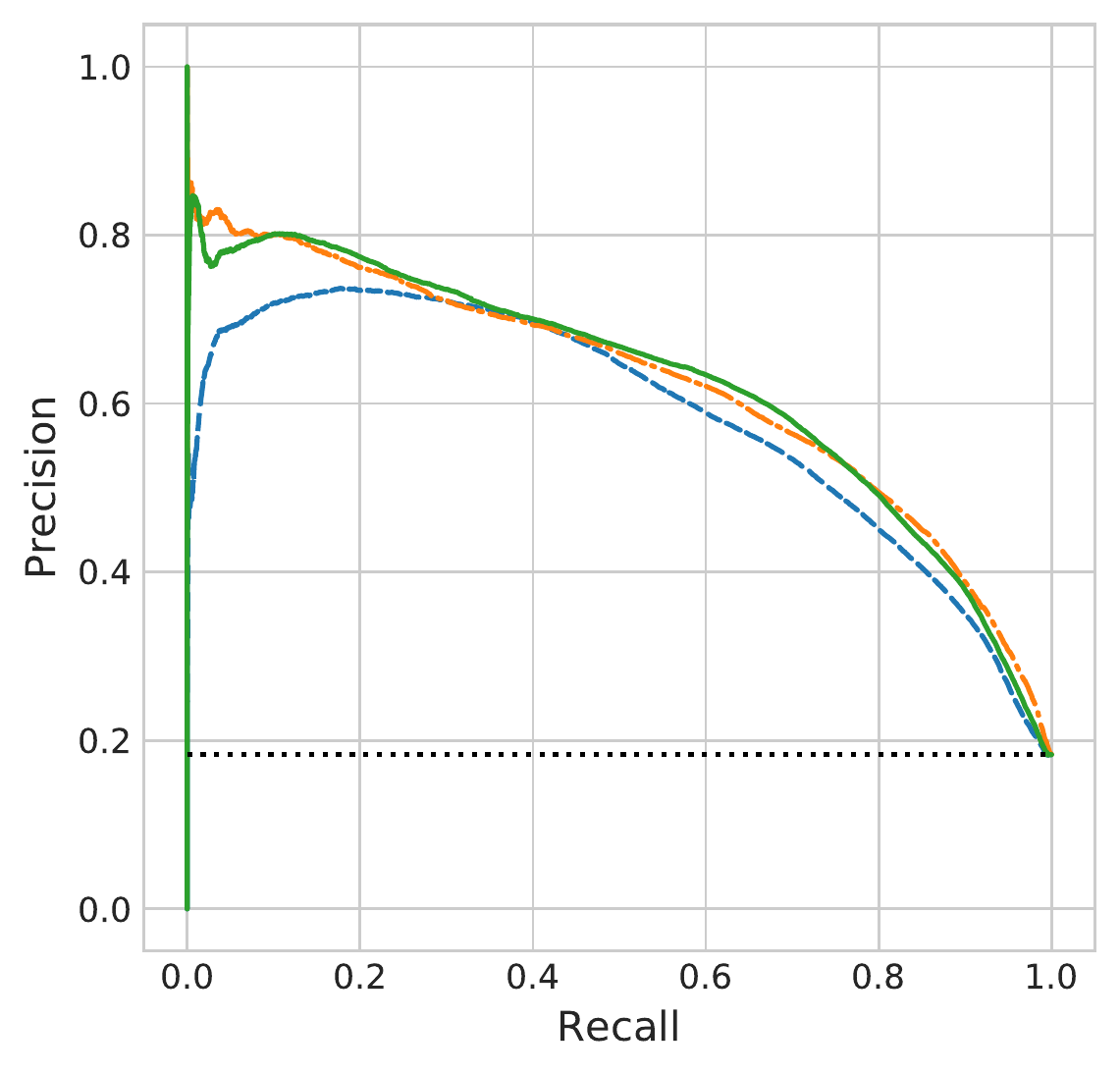}
      \caption{PR curve}
      \label{fig:precision_recall_curve}
	\end{subfigure}
    \caption{Performance curves for three different machine learning approaches: Logistic Regression, Gradient Boosting, and Feedforward NN; compared with the input-agnostic method that always predicts the major class. As the curves for Gradient Boosting and Feedforward NN lie higher than the curves for Logistic regression, we conclude that the corresponding models are better.}
    \label{fig:quality_comparison_curves}
\end{figure}

Figures \ref{fig:quality_comparison_curves} present visual comparison of performance of different classification methods.

\begin{figure}[H]
\centering
    \includegraphics[width=0.6\textwidth]{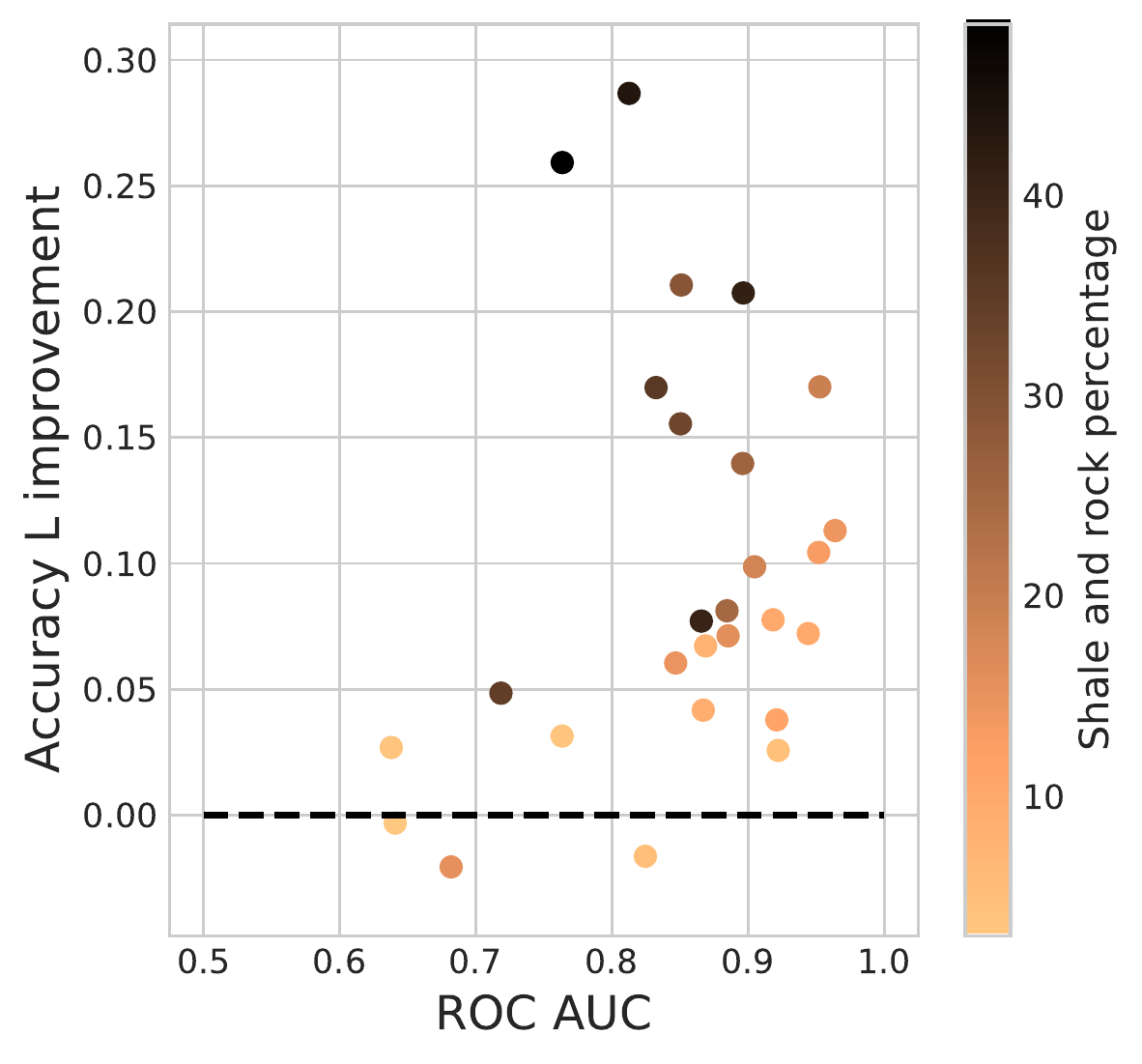}
    \caption{Gradient Boosting performance on different wells with respect to well-specific shale and rock percentage. The vertical axis represents the improvement of Accuracy L from using Gradient Boosting over the major class predictions.}
    \label{fig:roc_auc_accl_imp}
\end{figure}

Figure \ref{fig:roc_auc_accl_imp} shows performance of the Gradient Boosting with respect to lithotype classes balance. 
The lithotype predictions with the trained classifier are better than major-class predictions for 24 out of 27 wells. 
Improvement of Accuracy L increases if the classes are more balanced, that is, if they tend to have more equal shares of shales and rocks (first class), and sands (second class). 
However, the improvement varies significantly from well to well. Figure \ref{fig:classification_examples} shows examples of lithotype classification on three wells with different achieved quality. 

\definecolor{sand}{RGB}{255, 255, 187}
\definecolor{shale}{RGB}{142, 142, 139}
\begin{figure}
\centering
	\begin{subfigure}[b]{.32\linewidth}
      \includegraphics[width=0.99\textwidth]{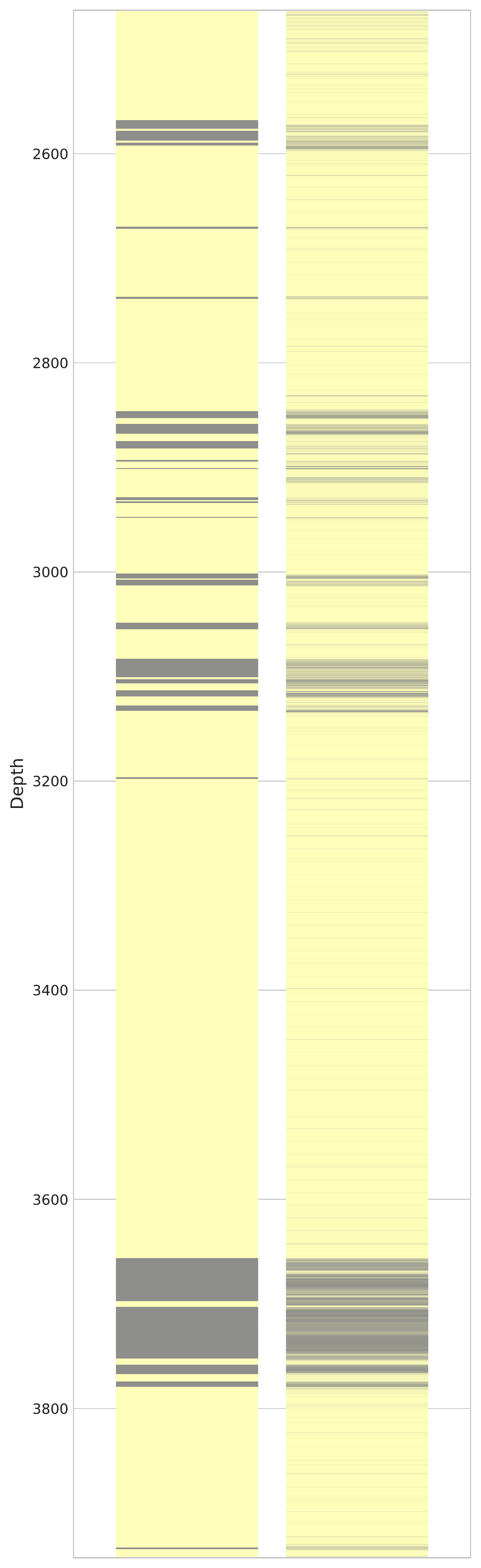}
      \caption{ROC AUC 0.9657}
    \end{subfigure}
    \begin{subfigure}[b]{.32\linewidth}
      \includegraphics[width=0.99\textwidth]{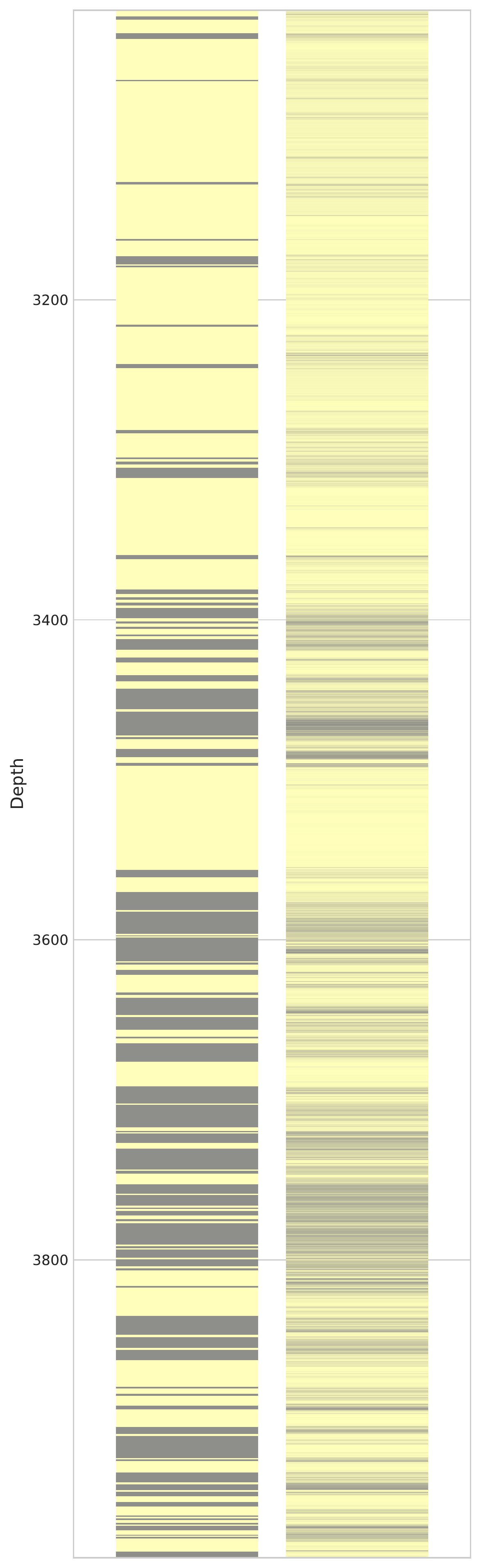}
      \caption{ROC AUC 0.8004}
    \end{subfigure}
    \begin{subfigure}[b]{.32\linewidth}
      \includegraphics[width=0.99\textwidth]{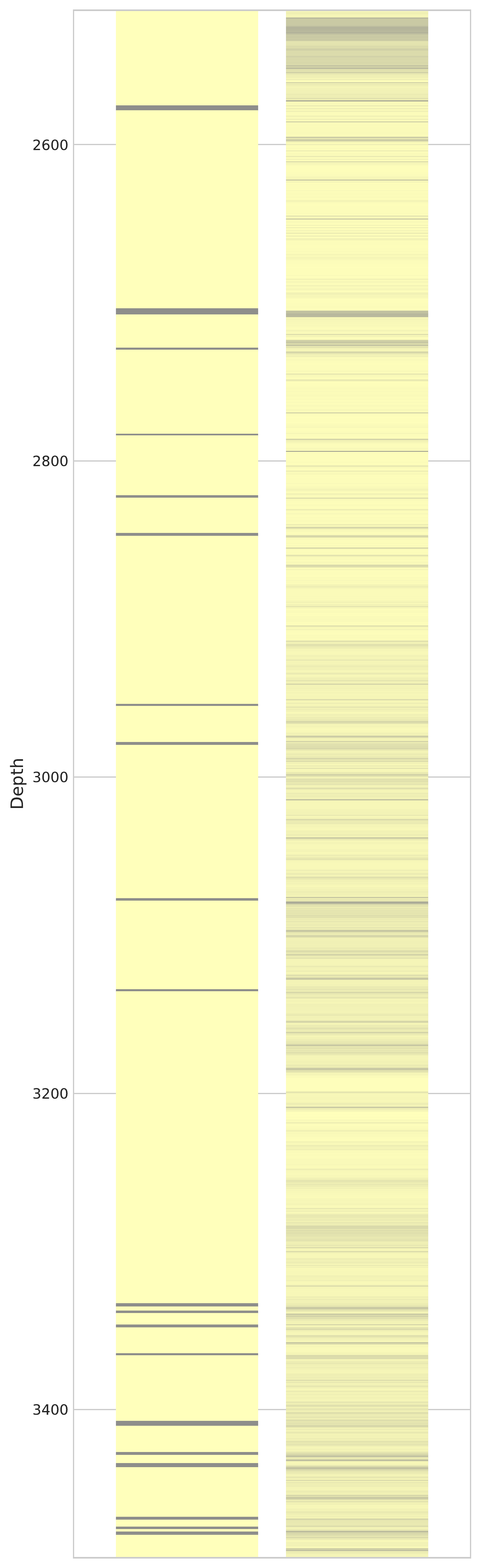}
      \caption{ROC AUC 0.6298}
    \end{subfigure}
    {
    \setlength{\fboxsep}{0pt}%
	\setlength{\fboxrule}{0.5pt}%
      \begin{tabular}{cc}
			
        \fbox{$\vcenter{\color{sand}\hbox{\rule{10pt}{5pt}}}$} sand & \fbox{$\vcenter{\color{shale}\hbox{\rule{10pt}{5pt}}}$}  shale or hard-rock\\

      \end{tabular}
 	}
    \caption{Examples of lithotype classification for three wells with different achieved quality: from one of the best on the left through average in the middle to one of the worst on the right. In each subfigure the left column shows the true lithotype values: yellow color represents sand, grey color represents shales and hard-rock; the right column shows the respective probability of lithotypes given by the classifier.}
    \label{fig:classification_examples}

\end{figure}

\section{Discussion}
\begin{figure}[H]
\centering
\includegraphics[width=0.9\textwidth]{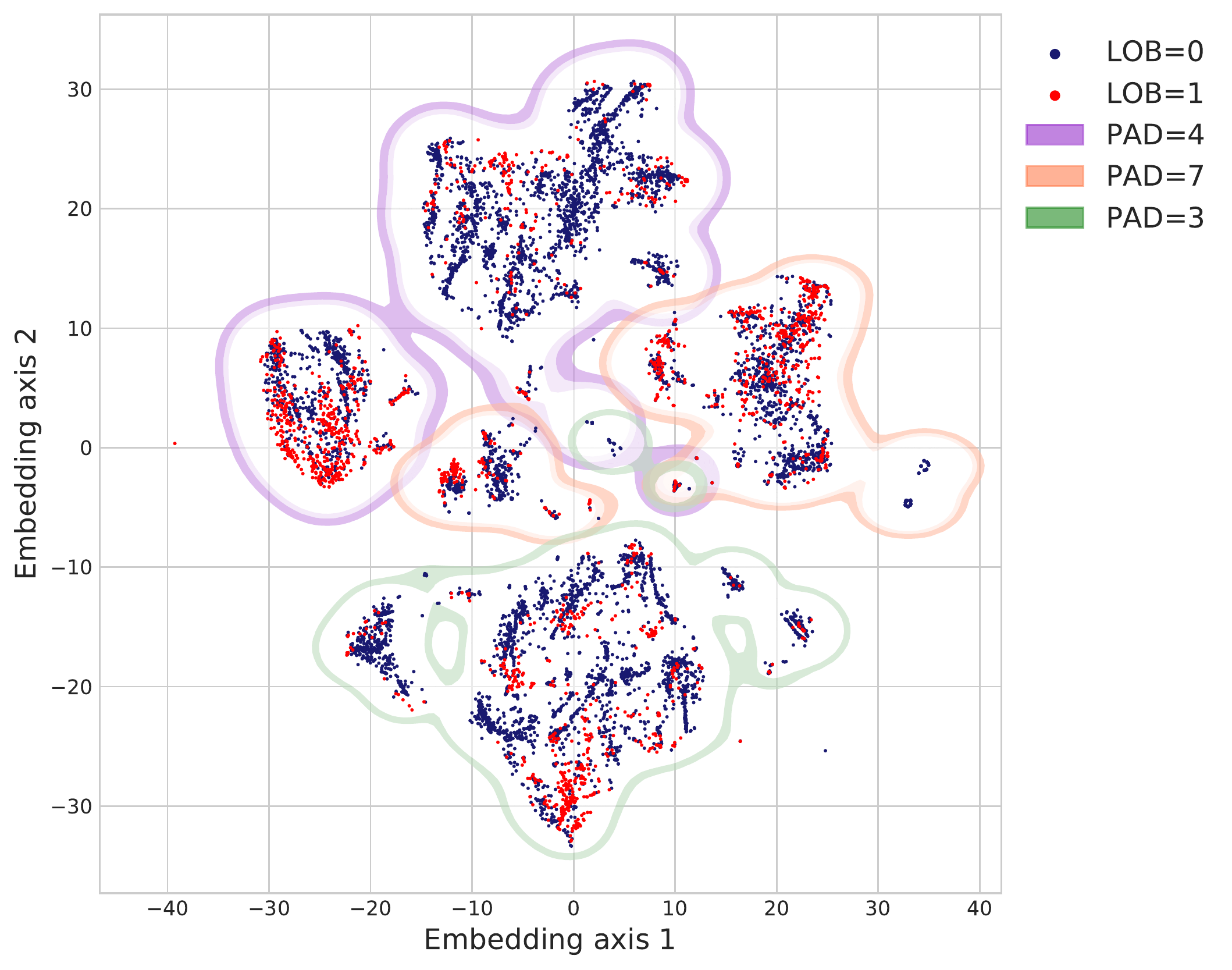}
\caption{2-dimensional embedding of the MWD data. Colors of scattered points indicate rock types in the corresponding drilling states. Contours indicate different PADs. It is easy to distinguish different PADs for this 2-dimensional embedding, while it is hard to distinguish two LOBs.}
\label{fig:phase_portrait}
\end{figure}

In this section, we discuss possible ways for improving the classification accuracy of the data-driven models. For this purpose, we study peculiarities of the initial data by embedding multi-dimensional MWD features in a convenient for analysis 2D space. In Figure \ref{fig:phase_portrait} we represent data applying a t-distributed Stochastic Neighborhood Embedding \citep{maaten2008visualizing} method on vectors of Basic and Lagged features including \APR{}, \SED{}, and their lagged values. 

The 2D representation shows the non-homogeneous nature of the real-world MWD measurements. In terms of Machine learning, this means that the algorithms are trained on a few localized areas of the data points, which dilutes the information among them. That is, we face a case when the algorithms make use of multiple small datasets instead of a single uniform large one. For example, we did not use features that explicitly specify pads, while the 2D representation of data has separated pads. Such distribution of data can negatively affect the generalization ability of classifiers, especially the ones that are based on threshold rules. 
Moreover, the mixture of different rock types and indistinct margins of classes illustrate fundamental indiscriminability of some part of data within the considered features.

One way to improve generalization ability is to use more discriminative features from additional sensors. Another way is to apply domain adaptation approach \citep{ganin2015unsupervised} for transforming input features for non-Neural Network algorithms like Gradient Boosting. However, the performance of Neural Networks is unlikely to be improved much with this approach, since they are capable of learning universal representations  \citep{bilen2017universal}.

Other ways for improving classification quality belong to three major areas.

The first area is consideration of different types of income data like LWD data, information about a well or a bit in total or drill cuttings. The main problem here is how to integrate different data sources of variable degree of fidelity and spatial resolution~\citep{zaytsev2017large} as the current approaches are often problem-specific especially when dealing with more than two levels of fidelity of data ~\citep{zaytsev2016reliable}.

The second area is related to correcting sample labels. One may want to use raw LWD data to train at and to predict, because LWD data will allow one to replace subjective lithotype interpretation made by experts with automatic labeling based on images at a training set markup, and will likely open new horizons for better resolution of the predictive model. 

The third area is the multi-class classification which is likely to allow distinguishing between several rock types rather than only a target interval and a boundary shale-reach zone. This will enrich the application of such data-driven predictions and move them from the point of just operative trajectory correction towards the capabilities to optimal control of the penetration rate with respect to maximal drilling efficiency at minimal tolerance to potential failures related to geomechanical specifics of the rocks. 

\section{Conclusion}
This study illustrates the capabilities of machine learning to handle the real technological issues of directional drilling. The accuracy of prediction of rock types relevant to directional drilling management reaches 91\%, that is, the classification error drops from 13.5\% (major-class prediction) down to 9\% (the best-achieved performance by examined algorithms). The involved algorithms allow real-time implementation which makes them useful for drilling support IT infrastructure. Further development of the predictive algorithms is covered in the discussion. 

\section*{References}

\bibliography{mybibfile}

\end{document}